\pdfoutput=1

\documentclass[11pt]{article}

\usepackage[final]{acl}

\usepackage{times}
\usepackage{latexsym}

\usepackage[T1]{fontenc}

\usepackage[utf8]{inputenc}

\usepackage{microtype}

\usepackage{inconsolata}

\usepackage{amsmath}
\usepackage{bm}
\usepackage{algpseudocode}
\usepackage[noend,ruled,linesnumbered]{algorithm2e}
\usepackage{color}
\usepackage{enumitem}
\usepackage{url}
\usepackage{graphicx}
\usepackage{array}
\usepackage{multirow}
\usepackage{pifont}
\usepackage{xcolor,colortbl}
\usepackage{setspace}
\usepackage{makecell}
\usepackage{balance}
\usepackage{booktabs} 
\usepackage{amsthm}
\usepackage{amsfonts}
\usepackage{subcaption}
\usepackage{mdframed}

\newtheorem{theorem}{Theorem}
\newtheorem{definition}[theorem]{Definition}

\newcommand{\ours}{\textsc{TaxoInstruct}\xspace}

\definecolor{myblue}{rgb}{0.2, 0.2, 0.9}
\definecolor{myorange}{rgb}{0.9, 0.5, 0.0}

\title{A Unified Taxonomy-Guided Instruction Tuning Framework for \\ Entity Set Expansion and Taxonomy Expansion}

\author{
Yanzhen Shen\textsuperscript{1},
Yu Zhang\textsuperscript{2},
Yunyi Zhang\textsuperscript{1},
Jiawei Han\textsuperscript{1} \\
\textsuperscript{1}University of Illinois Urbana-Champaign, 
\textsuperscript{2}Texas A\&M University \\
\texttt{\{yanzhen4,yzhan238,hanj\}@illinois.edu} \\
\texttt{yuzhang@tamu.edu}
}

\begin{document}

\maketitle

\begin{abstract}
Entity set expansion, taxonomy expansion, and seed-guided taxonomy construction are three representative tasks for automatically enriching an existing taxonomy with emerging concepts. 
Previous studies have treated them as separate tasks, leading to techniques that are specialized for one task but lack generalizability and a holistic perspective. 
In this paper, we propose a unified solution to address all three tasks.
Specifically, we identify two fundamental skills facilitating the three tasks: finding ``siblings'' and finding ``parents''.
To this end, we introduce a taxonomy-guided instruction tuning framework that trains a large language model to generate siblings and parents for query entities, where the joint pre-training process enables mutual reinforcement of these two skills.
Extensive experiments on multiple benchmark datasets validate the effectiveness of our proposed \ours framework, demonstrating its superiority over task-specific baselines across all three tasks. Our codes and data are available at \url{https://github.com/yanzhen4/TaxoInstruct}.
\end{abstract}

\section{Introduction}
Entities are fundamental to natural language processing. To better capture their semantics, taxonomies are constructed across various domains, including science~\cite{shen2018web}, e-commerce~\cite{mao2020octet}, and social media~\cite{gonccalves2019use}, to characterize the parent-child relationship between entities. While taxonomies are often initially curated by domain experts, the continuous emergence of new concepts necessitates automatic expansion to maintain their freshness and completeness. To this end, previous studies have explored three key tasks for integrating new entities into existing knowledge. 

\vspace{1mm}
\noindent \textbf{(1) Entity Set Expansion}~\cite{wang2007language,rong2016egoset,shen2017setexpan}: Given a set of entities belonging to a specific semantic class, the goal is to identify more entities within the same class. For example, given the seed entities $\{$\textit{Database}, \textit{Information Retrieval}, \textit{Operating System}$\}$, an entity set expansion algorithm should retrieve other computer science subfields such as \textit{Data Mining} and \textit{Human-Computer Interaction}. From a taxonomy perspective, this task can be viewed as finding ``\textbf{siblings}'' of existing entities.

\vspace{1mm}
\noindent \textbf{(2) Taxonomy Expansion}~\cite{shen2020taxoexpan,yu2020steam,zeng2021enhancing}: The goal of this task is to insert a provided new entity into an existing taxonomy by identifying its most appropriate ``\textbf{parents}''. For instance, consider a taxonomy with the root node \textit{Scientific Fields} and its children \textit{Computer Science}, \textit{Mathematics}, \textit{Physics}, and \textit{Chemistry}. Given a new concept \textit{Data Mining}, a taxonomy expansion model should place it as a child of \textit{Computer Science}.

\vspace{1mm}
\noindent \textbf{(3) Seed-Guided Taxonomy Construction}~\cite{shen2018hiexpan}: Given a seed taxonomy with a small number of entities, the goal is to construct a more comprehensive taxonomy that expands upon the initial structure. For example, if the input consists of \textit{Computer Science}, \textit{Chemistry}, and several of their subfields (e.g., \textit{Data Mining} and \textit{Organic Chemistry}), the expected output should include more scientific fields (e.g., \textit{Mathematics} and \textit{Physics}) and their subfields (e.g., \textit{Database}, \textit{Algebra}, and \textit{Astrophysics}), with explicitly identified parent-child relationships.
To approach this problem, we can first discover new entities at each layer and then figure out the parent-child edges between adjacent layers. Essentially, this can be framed as pipelining the steps of finding ``\textbf{siblings}'' and finding ``\textbf{parents}''.

\begin{figure*}[t]
\centering
\includegraphics[width=\linewidth]{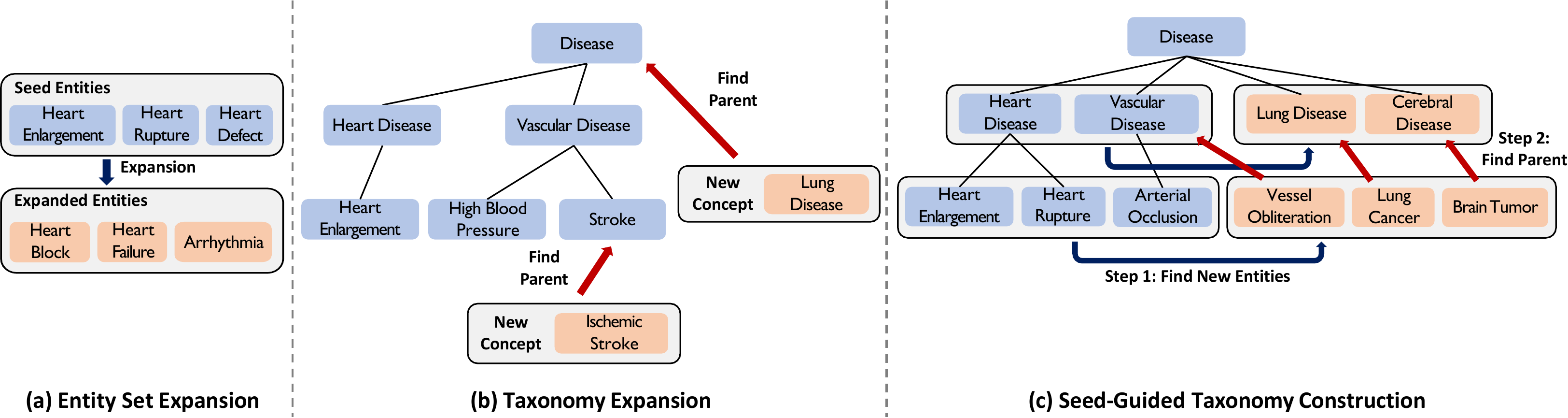}
\vspace{-1em}
\caption{Illustrations of the three tasks.}
\vspace{-1em}
\label{fig:taskdef}
\end{figure*}

As evident from the discussion above, all three tasks can be cast as finding entities that share a specific type of relationship with the given entities: entity set expansion involves finding ``siblings''; taxonomy expansion relies on finding ``parents''; seed-guided taxonomy construction integrates both. 
However, existing studies typically address only one of the three tasks, proposing task-specific techniques with little attention to their underlying commonalities. Intuitively, the processes of finding ``siblings'' and ``parents'' can reinforce each other. For example, recognizing that \textit{Data Mining} is a sibling of \textit{Database} and \textit{Information Retrieval} can help predict its parent as \textit{Computer Science}, and vice versa. By improving the accuracy of both sibling and parent prediction, we can leverage them as fundamental building blocks to solve all three tasks in a more holistic and unified manner.

\vspace{1mm}
\noindent \textbf{Contributions.} Building on the insights above, this paper proposes a unified framework to simultaneously address entity set expansion, taxonomy expansion, and seed-guided taxonomy construction. Specifically, we leverage existing taxonomies as rich sources of sibling-sibling and parent-child relationships to pre-train a model for identifying both siblings and parents. This pre-trained model can then be fine-tuned on domain-specific data (e.g., parent-child pairs from the input taxonomy in the taxonomy expansion task) to perform downstream tasks effectively. To implement this framework, we harness the instruction-following capabilities of large language models (LLMs)~\cite{wei2022finetuned,ouyang2022training}. Our proposed \ours framework employs task-specific instructions to train an LLM to generate sibling entities and identify parent entities for one or more query entities. The joint pre-training process enables mutual enhancement between these two skills and benefits overall performance of all three tasks.

To evaluate \ours, we conduct extensive experiments on 6 benchmark datasets spanning entity set expansion, taxonomy expansion, and seed-guided taxonomy construction. The results demonstrate that \ours, as a unified framework, significantly outperforms strong task-specific baselines across all three tasks. Additionally, we examine the impact of different LLM backbones~\cite{touvron2023llama,jiang2023mistral,team2024gemma} within \ours, showing that its effectiveness is robust and does not depend on a specific LLM choice.
\section{Task Definition}
\label{sec:task}

In this section, we formally introduce the three representative tasks for populating a taxonomy with new entities.

\vspace{1mm}
\noindent \textbf{Entity Set Expansion.} As shown in Figure \ref{fig:taskdef}(a), the entity set expansion task seeks to identify a set of "sibling" entities that belong to the same semantic class as a few example entities (referred to as "seeds"). Formally,

\begin{definition}{(Entity Set Expansion)}
Given a small set of seed entities $\mathcal{S}=\{s_1, s_2, ..., s_M\}$, the task is to discover more entities $\mathcal{S}^+=\{s_{M+1}, s_{M+2},$ $..., s_{M+N}\}$, where $s_1, s_2, ..., s_{M+N}$ fall into the same semantic category.
\label{def:setexpan}
\end{definition}

\vspace{1mm}
\noindent \textbf{Taxonomy Expansion.} As shown in Figure \ref{fig:taskdef}(b), taxonomy expansion involves inserting a set of new entities into an existing taxonomy by identifying an appropriate ``parent'' node in the taxonomy for each new entity. Formally,

\begin{definition}{(Taxonomy Expansion)}
Given an existing taxonomy $\mathcal{T}$ (which contains a set of entities $\mathcal{S}$ and the parent-child relationship between the entities $\textsc{Parent}(\cdot): \mathcal{S} \rightarrow \mathcal{S} \cup \{s_{\rm ROOT}\}$) and a set of new entities $\mathcal{S}^+$, the task is to expand the taxonomy to a more complete one $\mathcal{T}^+$ with entities $\mathcal{S} \cup \mathcal{S}^+$ and the parent-child relationship $\textsc{Parent}^+(\cdot): \mathcal{S} \cup \mathcal{S}^+ \rightarrow \mathcal{S} \cup \{s_{\rm ROOT}\}$.
\label{def:taxoexpan}
\end{definition}

\begin{figure*}[t]
\centering
\includegraphics[width=\linewidth]{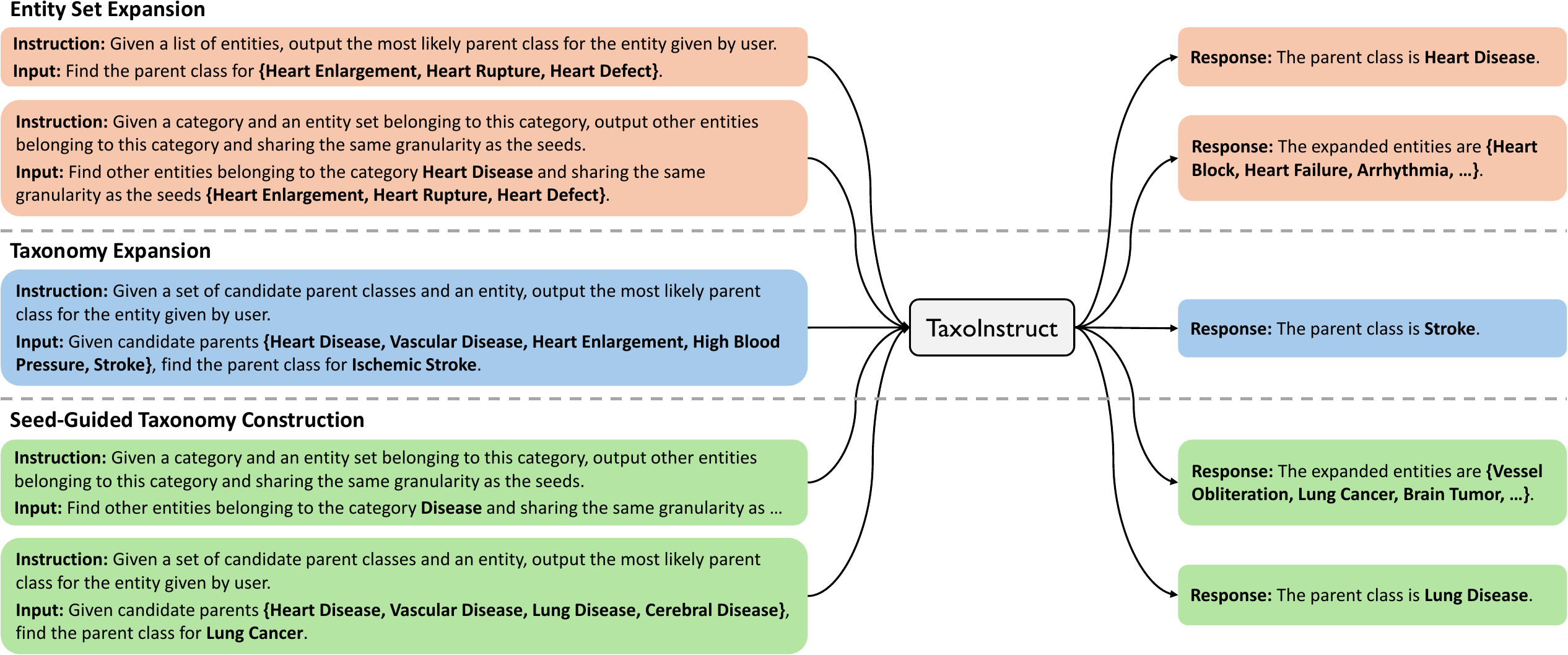}
\vspace{-1em}
\caption{Illustration of the \ours framework.}
\vspace{-1em}
\label{fig:model}
\end{figure*}

\vspace{1mm}
\noindent \textbf{Seed-Guided Taxonomy Construction.} As shown in Figure \ref{fig:taskdef}(c), seed-guided taxonomy construction involves two steps: first, identifying a set of new entities to be added to the taxonomy, and then determining the appropriate parent for each new entity.  

\begin{definition}{(Seed-Guided Taxonomy Construction)}
Given a small set of seeds that form a tree structure $\mathcal{T}=(\mathcal{S}_0, \mathcal{S}_1, ..., \mathcal{S}_L)$, where $\mathcal{S}_0$ is $\{s_{\rm ROOT}\}$, $\mathcal{S}_l$ ($1 \leq l \leq L$) denotes the set of seeds at layer $l$, and the parent-child relationship is characterized by a mapping function $\textsc{Parent}(\cdot): \mathcal{S}_l \rightarrow \mathcal{S}_{l-1}$, the task aims to discover more entities at each level (denoted by the sets $\mathcal{S}_1^+, ..., \mathcal{S}_L^+$, where entities in $\mathcal{S}_l$ and $\mathcal{S}_l^+$ belong to the same semantic class) and predict their parent-child relationship (characterized by $\textsc{Parent}^+(\cdot): \mathcal{S}_l \cup \mathcal{S}_l^+ \rightarrow \mathcal{S}_{l-1} \cup \mathcal{S}_{l-1}^+$).
\label{def:hiexpan}
\end{definition}


\section{Model}
\label{sec:model}

Inspired by the intuition that entity set expansion, taxonomy expansion, and seed-guided taxonomy construction all rely on two fundamental skills—finding ``siblings'' and finding ``parents''—we aim to train a unified model that simultaneously learns both skills, thereby facilitating all three tasks. To implement this idea, in this section, we propose \ours, a unified taxonomy-guided instruction tuning framework.

\subsection{Entity Set Expansion} \label{sec:setexpan}

Given a set of seeds $\mathcal{S}=\{s_1, s_2, ..., s_M\}$, the entity set expansion task imposes two constraints on the expanded entities $\mathcal{S}^+=\{s_{M+1}, s_{M+2},$ $..., s_{M+N}\}$.
First, $s_{M+n}$ $(1 \leq n \leq N)$ must belong to the same semantic category as $s_1, s_2, ...,$ $s_M$. For example, in Figure \ref{fig:taskdef}(a), both \textit{Heart Enlargement} and \textit{Arrhythmia} belong to the category \textit{Heart Disease}. Second, $s_{M+n}$ must share the same level of granularity as $s_1, s_2, ..., s_M$. For example, while \textit{Congenital Heart Defect} also belongs to \textit{Heart Disease}, it should not be expanded in Figure \ref{fig:taskdef}(a) because it is more fine-grained than the seed \textit{Heart Defect}. These two restrictions inherently describe the concept of ``siblings'' in a taxonomy, as siblings share the same parent and reside at the same hierarchical level.  

Inspired by this, we frame the entity set expansion task (from a taxonomy perspective) as identifying other siblings of the seed entities. We tackle this problem by leveraging the ability of LLMs to follow task-specific instructions~\cite{wei2022finetuned,ouyang2022training}. Briefly, given a set of \textsc{Input} entities $\mathcal{S}=\{s_1, s_2, ..., s_M\}$ that share the same parent node $\textsc{Parent}(\mathcal{S})$, we \textsc{Instruct} an LLM (e.g., 
Llama-3 8B~\cite{dubey2024llama}) to generate more children of $\textsc{Parent}(\mathcal{S})$ in its \textsc{Response}.

Nevertheless, the parent entity $\textsc{Parent}(\mathcal{S})$ is not available in the standard entity set expansion task~\cite{rong2016egoset,shen2017setexpan}. Thus, we first prompt the LLM to generate the parent entity for the seed set $\mathcal{S}$. Following the (\textsc{Instruction}, \textsc{Input}, \textsc{Response}) schema of Llama-3, we form the instruction as follows: 

\vspace{1mm}
\begin{mdframed}
\small
\textsc{Instruction}: \textit{Given a list of entities, output the most likely parent class for the entity given by user.} 

\vspace{1mm}
\noindent \textsc{Input}: \textit{Find the parent class for} $\{s_1, s_2, ..., s_M\}$.

\vspace{1mm}
\noindent \textsc{Response}: \textit{The parent class is} 
\end{mdframed}

\vspace{1mm}
The generated parent entity $\textsc{Parent}(\mathcal{S})$ is then used to guide the expansion process:

\vspace{1mm}
\begin{mdframed}
\small
\textsc{Instruction}: \textit{Given a category and an entity set belonging to this category, output other entities belonging to this category and sharing the same granularity as the seeds.}

\vspace{1mm}
\noindent \textsc{Input}: \textit{Find other entities belonging to the category} $\textsc{Parent}(\mathcal{S})$ \textit{and sharing the same granularity as the seeds} $\{s_1, s_2, ..., s_M\}$.

\vspace{1mm}
\noindent \textsc{Response}: \textit{The expanded entities are}
\end{mdframed}

\vspace{1mm}
The LLM will generate a set of expanded entities, which we denote as $\mathcal{R}=\{r_1, r_2, ..., r_K\}$. 
After that, we perform a ranking step to sort these entities. To be specific, we use a pre-trained encoder language model (e.g., BERT~\cite{devlin2019bert}) to compute the similarity score between each generated entity $r \in \mathcal{R}$ and $\textsc{Parent}(\mathcal{S})$: 
\begin{equation}
\small
    {\rm sim}(r, \textsc{Parent}(\mathcal{S})) = \cos\Big({\rm E}(r), {\rm E}(\textsc{Parent}(\mathcal{S}))\Big),
\label{eq:post-sibling}
\end{equation}
where ${\rm E}(\cdot)$ denotes the average output token embedding after feeding the entity name into the pre-trained encoder. All entities in $\mathcal{R}$ are then ranked according to ${\rm sim}(\cdot, \textsc{Parent}(\mathcal{S}))$. Afterwards, we add the top-ranked entities to the seed entity set $\mathcal{S}$ and rerun the expansion process with the enriched seed set. This process can be conducted iteratively, following the common practice of previous entity set expansion algorithms~\cite{shen2017setexpan,zhang2020empower}.
After the final iteration, we rank all seeds and expanded entities (except the original seeds which should not appear in the output) according to ${\rm sim}(\cdot, \textsc{Parent}(\mathcal{S}))$ and obtain a list, $\mathcal{S}^{+}$, of expanded entities.

\subsection{Taxonomy Expansion} \label{sec:taxoexpan}

Taxonomy expansion is a parent-finding task. Given an \textsc{Input} entity $s_q \in \mathcal{S}^+$, we \textsc{Instruct} an LLM to identify the correct parent node $\textsc{Parent}(s_q)$ from a provided list of candidates $\mathcal{S}=\{s_1, s_2, ..., s_M\}$ (i.e., entities in the existing taxonomy): 

\vspace{1mm}
\begin{mdframed}
\small
\textsc{Instruction}: \textit{Given a set of candidate parent classes and an entity, output the most likely parent class for the entity given by user.}

\vspace{1mm}
\noindent \textsc{Input}: \textit{Given candidate parents} $\{s_1, s_2, ..., s_M\}$, \textit{find the parent class for} $s_q$.

\vspace{1mm}
\noindent \textsc{Response}: \textit{The parent class is}
\end{mdframed}

\vspace{1mm}
In practice, however, the input taxonomy may contain a large number of (e.g., more than 10,000) entities~\cite{shen2020taxoexpan}. If we include all of them as candidates and put them into the instruction, the LLM may be overwhelmed by the overly large label space and can hardly follow the instruction. To tackle this problem, we first retrieve a set of candidates from the taxonomy and thus reduce the label space for the LLM. More specifically, given the query $s_q$, we select top-$U$ (e.g., $U = 20$) entities $\mathcal{U}_q \subseteq \mathcal{S}$ with the highest similarity to $s_q$.

\begin{equation}
\small
\mathcal{U}_q = \arg\max_{\mathcal{U} \subseteq \mathcal{S}, |\mathcal{U}|=U} \sum_{s \in \mathcal{U}} \cos\Big({\rm E}(s_q), {\rm E}(s)\Big).
\label{eq:candidate-parent}
\end{equation}
The retrieved subset $\mathcal{U}_q$ will replace the entire candidate list in the \textsc{Input}.

Since the input taxonomy contains a wealth of (parent, child) entity pairs, we leverage this information to fine-tune the LLM, enhancing its understanding of parent-child relationships and domain-specific knowledge. To be specific, given a node $s_i$ in the input taxonomy and its parent $\textsc{Parent}(s_i)$, we construct fine-tuning data in two different ways.

First, we take the siblings of $\textsc{Parent}(s_i)$ as distractors. In other words, the LLM needs to identify the true parent $\textsc{Parent}(s_i)$ from the candidates $\{\textsc{Parent}(s_i)\} \cup \textsc{Sibling}(\textsc{Parent}(s_i))$.

Second, we use Eq. (\ref{eq:candidate-parent}) to find the set of top-$U$ entities $\mathcal{U}_i$ that are closest to $s_i$. Then, the LLM needs to identify the true parent $\textsc{Parent}(s_i)$ from the candidates $\{\textsc{Parent}(s_i)\} \cup \mathcal{U}_i$.

Filling $s_i$ and the candidates into our instruction template, we fine-tune the LLM to generate $\textsc{Parent}(s_i)$.




\subsection{Seed-Guided Taxonomy Construction}
\label{sec:hiexpan}

As shown in Figure \ref{fig:taskdef}(c), seed-guided taxonomy construction can be naturally divided into two subtasks: (1) expanding the entity set at each layer to discover new entities (i.e., finding ``siblings'' and ``cousins''\footnote{In the first step of seed-guided taxonomy construction, the goal is to find entities that share the same semantic granularity as the seeds at each layer. These entities are required only to be descendants of the root node and may not necessarily share the same parent as the seeds. Therefore, this step involves discovering not just ``siblings'' but also ``cousins''.}) and (2) expanding the taxonomy by specifying the proper ``parent'' for each new entity. 
Since these two subtasks closely align with entity set expansion and taxonomy expansion, respectively, we can leverage similar instructions as outlined in Sections \ref{sec:setexpan} and \ref{sec:taxoexpan}.

\vspace{1mm}
\noindent \textbf{Finding ``Siblings'' and ``Cousins''.} Given the input taxonomy $\mathcal{T}=(\mathcal{S}_0, \mathcal{S}_1, ..., \mathcal{S}_L)$ where $\mathcal{S}_0 = \{s_{\rm ROOT}\}$ and $\mathcal{S}_l=\{s_{l,1}, s_{l,2}, ...,$ $s_{l,M_l}\}$ $(1\leq l \leq L)$, we adopt the following instruction:

\vspace{1mm}
\begin{mdframed}
\small
\textsc{Instruction}: \textit{Given a category and an entity set belonging to this category, output other entities belonging to this category and sharing the same granularity as the seeds.}

\vspace{1mm}
\noindent \textsc{Input}: \textit{Find other entities belonging to the category} $s_{\rm ROOT}$ \textit{and sharing the same granularity as the seeds} $\{s_{l,1}, s_{l,2}, ..., s_{l,M_l}\}$.

\vspace{1mm}
\noindent \textsc{Response}: \textit{The expanded entities are}
\end{mdframed}

\vspace{1mm}
The major difference between this instruction and that for entity set expansion is that we put $s_{\rm ROOT}$ rather than $\textsc{Parent}(\mathcal{S}_l)$ into the \textsc{Input} to discover not only ``siblings'' but also ``cousins'' of $\mathcal{S}_l$.
We denote the expanded entities at layer $l$ as $\mathcal{S}_l^+=\{s_{l,M_l+1}, s_{l,M_l+2}, ...,$ $s_{l,M_l+N_l}\}$ $(1\leq l \leq L)$. 


\vspace{1mm}
\noindent \textbf{Finding ``Parents''.} For each newly discovered entity $s_{l,M_{l}+n} \in \mathcal{S}_l^+ \backslash \mathcal{S}_l$, we need to insert it into the taxonomy by finding its parent from all entities that are one layer coarser. When $l=1$, this problem is trivial because the parent is $s_{\rm ROOT}$. When $l \geq 2$, we consider the following instruction:

\vspace{1mm}
\begin{mdframed}
\small
\textsc{Instruction}: \textit{Given a set of candidate parent classes and an entity, output the most likely parent class for the entity given by user.}

\vspace{1mm}
\noindent \textsc{Input}: \textit{Given candidate parents} $\{s_{l-1,1}, s_{l-1,2}, ...,$ $s_{l-1,M_{l-1}+N_{l-1}}\}$, \textit{find the parent for} $s_{l,M_{l}+n}$.

\vspace{1mm}
\noindent \textsc{Response}: \textit{The parent class is}
\end{mdframed}

\vspace{1mm}
The major difference between this instruction and that for taxonomy expansion is that the candidate parent list in the \textsc{Instruction} contains entities at layer $l-1$ only (i.e., $\mathcal{S}_{l-1}^+$) rather than the entire input taxonomy.

In seed-guided taxonomy construction, similar to taxonomy expansion, we are given a taxonomy structure $\mathcal{T}$ as input. Thus, we can also construct training data from $\mathcal{T}$ to fine-tune the LLM. Following Section~\ref{sec:taxoexpan}, for each seed $s_{l,m} \in \mathcal{S}_l$ $(l \geq 2)$, we train the LLM to pick the correct parent node $\textsc{Parent}(s_{l,m})$ from $\mathcal{S}_{l-1}$. 



\subsection{A Unified Pre-training Framework}
\label{sec:unified}

With the above instructions, an LLM can be directly prompted or fine-tuned to perform each task separately. However, task-specific training data may be too limited for the model to effectively learn the necessary skills for identifying siblings and parents. For instance, the input taxonomy for seed-guided taxonomy construction typically contains about 10 entities only~\cite{shen2018hiexpan}.
To address this limitation, we propose to first continuously pre-train a general-purpose LLM on a large existing taxonomy using the aforementioned instructions. This pre-training step allows the model to acquire broader knowledge and skills, which can then be transferred to the three tasks, enhancing its performance even with limited task-specific data.

\vspace{1mm}
\noindent \textbf{Pre-training Data.} To largely avoid overlap between pre-training data and evaluation benchmarks in downstream tasks (e.g., Wikipedia, SemEval, and DBLP), we adopt only one existing large-scale taxonomy for pre-training: Comparative Toxicogenomics Database (CTD)~\cite{davis2023comparative}, where we take its MEDIC taxonomy of disease entities.

\vspace{1mm}
\noindent \textbf{Pre-training Tasks.} 
Given a set of sibling entities $\mathcal{S}=\{s_1, s_2, ..., s_{|\mathcal{S}|}\}$ and their parent $\textsc{Parent}(\mathcal{S})$ in the taxonomy used for pre-training, we randomly pick $M$ entities from $\mathcal{S}$ as seeds. For ease of notation, we denote the seeds as $s_1, s_2, ..., s_M$. 

For the sibling-finding task, the pre-training objective is to generate $s_{M+1}, ..., s_{|\mathcal{S}|}$ from the seeds,
where the instruction follows the sibling-finding template in Section \ref{sec:setexpan}.
For the parent-finding task, the pre-training objective is to generate $\textsc{Parent}(\mathcal{S})$ for each individual seed $s_i$ $(1 \leq i \leq M)$ as well as for the entire set of seeds $\{s_1, s_2, ..., s_M\}$,
where the instruction follows the parent-finding template introduced in Section \ref{sec:taxoexpan}.
Intuitively, the two pre-training tasks mutually benefit each other because accurately predicting the siblings $s_{M+1}, ..., s_{|\mathcal{S}|}$ of $s_1, s_2, ..., s_M$ helps inferring the parent $\textsc{Parent}(\mathcal{S})$ of $s_1, s_2, ..., s_M$, and vice versa.

\section{Experiments}
We evaluate the effectiveness of \ours across all three tasks by comparing it with competitive baselines on benchmark datasets. Details of the baselines and evaluation metrics are provided in Appendices \ref{app:baseline} and \ref{app:metric}, respectively.

\subsection{Entity Set Expansion}
\noindent \textbf{Datasets.} Following previous studies~\cite{shen2017setexpan,yan2019learning,zhang2020empower}, we use two benchmark datasets, \textbf{APR} and \textbf{Wiki}, to evaluate entity set expansion algorithms. The two datasets are derived from news articles (published by Associated Press and Reuters) and Wikipedia articles, respectively.

\vspace{1mm}
\noindent \textbf{Baselines.}
The baselines for entity set expansion include 
\textbf{EgoSet} \cite{rong2016egoset}, 
\textbf{SetExpan} \cite{shen2017setexpan}, 
\textbf{SetExpander} \cite{mamou2018setexpander}, 
\textbf{CaSE} \cite{yu2019corpus}, 
\textbf{SetCoExpan} \cite{huang2020guiding}, 
\textbf{CGExpan} \cite{zhang2020empower}, 
\textbf{SynSetExpan} \cite{shen2020synsetexpan},  
\textbf{ProbExpan} \cite{li2022contrastive}, and
\textbf{Llama-3.1 70B} \cite{dubey2024llama}. 
Additionally, since \ours is pre-trained on both parent-finding and sibling-finding tasks, we investigate whether the former enhances the latter. To assess this, we introduce an ablation variant, \textbf{NoParentPretrain}, which is pre-trained on the sibling-finding task only.

\vspace{1mm}
\noindent \textbf{Evaluation Metric.}
Following previous studies~\cite{shen2017setexpan,yan2019learning,zhang2020empower}, we adopt the Mean Average Precision (\textbf{MAP@$k$}) as the evaluation metric.

\begin{table}[t]
\centering
\caption{Performance of compared methods in the entity set expansion task. \textbf{Bold}: the best score. *: \ours is significantly better than this method with p-value $<0.05$. $^\dagger$, $^\ddagger$, and $^\triangleright$: the scores of this method are reported in \citet{zhang2020empower}, \citet{huang2020guiding}, and \citet{li2022contrastive}, respectively.}
\vspace{-0.5em}
\resizebox{\linewidth}{!}{
\begin{tabular}{lllll}
\toprule
\multirow{2}{*}{\textbf{Method}} & \multicolumn{2}{c}{\textbf{APR}}  & \multicolumn{2}{c}{\textbf{Wiki}} \\
                                 & \textbf{MAP@10} & \textbf{MAP@20} & \textbf{MAP@10} & \textbf{MAP@20} \\
\midrule
EgoSet~$^\dagger$                           & 0.758$^*$            & 0.710$^*$            & 0.904$^*$            & 0.877$^*$            \\
SetExpan~$^\dagger$                         & 0.789$^*$            & 0.763$^*$            & 0.944$^*$            & 0.921$^*$            \\
SetExpander~$^\dagger$                      & 0.287$^*$            & 0.208$^*$            & 0.499$^*$            & 0.439$^*$            \\
CaSE~$^\dagger$                             & 0.619$^*$            & 0.494$^*$            & 0.897$^*$            & 0.806$^*$            \\
SetCoExpan~$^\ddagger$                       & 0.933$^*$            & 0.915$^*$            & 0.976$^*$            & 0.964$^*$            \\
CGExpan~$^\dagger$                          & 0.992            & 0.990$^*$            & 0.995            & 0.978$^*$            \\
SynSetExpan~$^\triangleright$                      & 0.985$^*$            & 0.990$^*$            & 0.991$^*$            & 0.978$^*$            \\
ProbExpan~$^\triangleright$                        & 0.993            & 0.990$^*$            & 0.995            & 0.982            \\
Llama-3.1 70B                        & 0.9933            & 0.9788$^*$            & 0.9861$^*$            & 0.9748$^*$            \\
\midrule
\ours                     & \textbf{0.9956}  & \textbf{0.9928}  & \textbf{0.9957}  & \textbf{0.9875} \\ 
\ \ NoParentPretrain & 0.9867$^*$ & 0.9689$^*$ & 0.9746$^*$ & 0.9720$^*$ \\
\bottomrule
\end{tabular}
}
\label{tab:setexpan}
\vspace{-1em}
\end{table}

\vspace{1mm}
\noindent \textbf{Implementation Details.}
We initialize our model with Llama-3 8B~\cite{dubey2024llama} and continuously pre-train/fine-tune it using Low-Rank Adaptation (LoRA)~\cite{hu2021lora}. 
The optimizer is AdamW~\cite{loshchilov2017decoupled}, and the batch size is 64.
We adopt SPECTER~\cite{cohan2020specter} as the pre-trained encoder ${\rm E}(\cdot)$ in Eqs. (\ref{eq:post-sibling}) and (\ref{eq:candidate-parent}). 

\vspace{1mm}
\noindent \textbf{Experimental Results.}
Table \ref{tab:setexpan} presents the ${\rm MAP}@10$ and $20$ scores of compared methods in the entity set expansion task. We run \ours multiple times and report the average performance. To assess statistical significance, we conduct a two-tailed Z-test comparing \ours against each baseline, with significance levels indicated in Table \ref{tab:setexpan}. We can observe that: 
(1) \ours consistently outperforms all baselines, including those leveraging language model probing (e.g., CGExpan and ProbExpan). In most cases, the advantage of \ours is statistically significant. 
(2) \ours performs significantly better than NoParentPretrain, suggesting that even in entity set expansion—where identifying siblings is the primarily required skill—pre-training \ours to find parents still enhances the performance. This finding validates our motivation for pre-training a unified model to jointly address different yet related tasks.

\subsection{Taxonomy Expansion}
\noindent \textbf{Datasets.} 
Following~\cite{jiang2023single}, we use two benchmark datasets, \textbf{Environment} and \textbf{Science}, from the shared task in SemEval 2016~\cite{bordea2016semeval}. Entities in these two datasets are scientific concepts related to environment and general science, respectively. 

\vspace{1mm}
\noindent \textbf{Baselines.}
The baselines for taxonomy expansion include 
\textbf{TAXI} \cite{panchenko2016taxi}, 
\textbf{HypeNET} \cite{shwartz2016improving}, 
\textbf{BERT+MLP} \cite{devlin2019bert}, 
\textbf{TaxoExpan} \cite{shen2020taxoexpan}, 
\textbf{Arborist} \cite{manzoor2020expanding}, 
\textbf{Graph2Taxo} \cite{shang2020taxonomy}, 
\textbf{STEAM} \cite{yu2020steam}, 
\textbf{TMN} \cite{zhang2021taxonomy}, 
\textbf{TEMP} \cite{liu2021temp}, 
\textbf{GenTaxo} \cite{zeng2021enhancing}
\textbf{BoxTaxo} \cite{jiang2023single}, and
\textbf{Llama-3.1 70B} \cite{dubey2024llama}. 
Additionally, to investigate if sibling finding helps parent finding, we introduce an ablation version of \ours, \textbf{NoSiblingPretrain}, for the taxonomy expansion task, which is pre-trained on the parent-finding task only.

\begin{table}[t]
\centering
\caption{Performance of compared methods in the taxonomy expansion task. \textbf{Bold} and *: the same meaning as in Table~\ref{tab:setexpan}. $^\dagger$, $^\ddagger$, and $^\triangleright$: the scores of this method are reported in \citet{jiang2023single}, \citet{zeng2021enhancing}, and \citet{liu2021temp}, respectively.}
\vspace{-0.5em}
\resizebox{0.92\linewidth}{!}{
\begin{tabular}{lllll}
\toprule
\multirow{2}{*}{\textbf{Method}} & \multicolumn{2}{c}{\textbf{Environment}} & \multicolumn{2}{c}{\textbf{Science}} \\
& \textbf{Acc}        & \textbf{Wu\&P}     & \textbf{Acc}      & \textbf{Wu\&P}   \\
\midrule
TAXI~$^\dagger$ & 0.167$^*$                & 0.447$^*$               & 0.130$^*$              & 0.329$^*$             \\
HypeNET~$^\dagger$ & 0.167$^*$                & 0.558$^*$               & 0.154$^*$              & 0.507$^*$             \\
BERT+MLP~$^\dagger$ & 0.111$^*$                & 0.479$^*$               & 0.115$^*$              & 0.436$^*$             \\
TaxoExpan~$^\dagger$ & 0.111$^*$                & 0.548$^*$               & 0.278$^*$              & 0.576$^*$             \\
Arborist~$^\ddagger$ & 0.4615$^*$ & -- & 0.4193$^*$ & -- \\
Graph2Taxo~$^\ddagger$ & 0.2105$^*$ & -- & 0.2619$^*$ & -- \\
STEAM~$^\dagger$ & 0.361$^*$                & 0.696$^*$               & 0.365$^*$              & 0.682$^*$             \\
TMN~$^\ddagger$ & 0.3793$^*$ & -- & 0.3415$^*$ & -- \\
TEMP~$^\triangleright$ & 0.492$^*$ & 0.777$^*$ & 0.578$^*$ & 0.853 \\
GenTaxo~$^\ddagger$ & 0.4828$^*$ & -- & 0.3878$^*$ & -- \\
BoxTaxo~$^\dagger$ & 0.381$^*$                & 0.754$^*$               & 0.318$^*$              & 0.647$^*$             \\
Llama-3.1 70B & 0.3654$^*$                & 0.6957$^*$               & 0.4471$^*$              & 0.7310$^*$             \\
\midrule
\ours & \textbf{0.5115}      & \textbf{0.8300}     & \textbf{0.6165}    & 0.8480 \\
\ \ NoSiblingPretrain & 0.4616$^*$ & 0.7911$^*$ & 0.5953$^*$ & \textbf{0.8559} \\
\bottomrule
\end{tabular}
}
\label{tab:taxoexpan}
\vspace{-1em}
\end{table}

\vspace{1mm}
\noindent \textbf{Evaluation Metrics.}
We adopt Accuracy (\textbf{Acc}) and Wu \& Palmer Similarity (\textbf{Wu\&P})~\cite{wu1994verbs} as the evaluation metrics.
Previous studies~\cite{yu2020steam,zeng2021enhancing,jiang2023single} also consider the mean reciprocal rank (MRR) as an evaluation metric. However, it requires a model to rank all nodes in the taxonomy according to their likelihood of being the parent, which is not applicable to \ours that generates only one predicted parent entity.

\vspace{1mm}
\noindent \textbf{Experimental Results.}
Table \ref{tab:taxoexpan} shows the performance of compared methods in taxonomy expansion. Our key observations are:
(1) \ours significantly outperforms all baselines in nearly every case. The only exception is that TEMP achieves a higher Wu\&P score on the Science dataset. Apart from TEMP, GenTaxo is a strong baseline that follows a generative paradigm for taxonomy expansion. However, unlike \ours, which leverages LLMs to fully harness the strengths of the generative approach, GenTaxo relies solely on a Gated Recurrent Unit (GRU) architecture, resulting in suboptimal performance.
(2) \ours outperforms NoSiblingPretrain across most columns, suggesting that even in the taxonomy expansion task—where identifying parent entities is the primary objective—pre-training the model to accurately identify siblings remains beneficial. Combined with our ablation analysis from entity set expansion, this finding supports the conclusion that sibling-finding and parent-finding skills can \textit{mutually} enhance each other.

\subsection{Seed-Guided Taxonomy Construction}
\noindent \textbf{Datasets.}
We use the \textbf{DBLP} and \textbf{PubMed-CVD} datasets introduced by~\citet{shen2018hiexpan}.
The seeds in our experiments are identical to those in~\citet{shen2018hiexpan}.
Both datasets have a two-layer input taxonomy.
For DBLP, there are 5 seeds at the top layer (i.e., \textit{Machine Learning}, \textit{Data Mining}, \textit{Natural Language Processing}, \textit{Information Retrieval}, and \textit{Wireless Networks}) and 11 seeds at the bottom layer.
For PubMed-CVD, there are 3 seeds at the top layer (i.e., \textit{Cardiovascular Abnormalities}, \textit{Vascular Diseases}, and \textit{Heart Disease}) and 10 seeds at the bottom layer.

\begin{table}[t]
\centering
\caption{Performance of compared methods in the seed-guided taxonomy construction task. \textbf{Bold} and *: the same meaning as in Table~\ref{tab:setexpan}.}
\vspace{-0.5em}
\resizebox{0.95\linewidth}{!}{
\begin{tabular}{lllll}
\toprule
\multirow{3}{*}{\textbf{Method}} & \multicolumn{2}{c}{\textbf{DBLP}} & \multicolumn{2}{c}{\textbf{PubMed-CVD}} \\
& \textbf{Sibling} & \textbf{Parent} & \textbf{Sibling} & \textbf{Parent} \\
& \textbf{nDCG} & \textbf{nDCG} & \textbf{nDCG} & \textbf{nDCG} \\
\midrule
HSetExpan & 0.8814$^*$ & 0.8268$^*$ & 0.6515$^*$ & 0.5085$^*$ \\
NoREPEL & 0.8830$^*$ & 0.8152$^*$ & 0.6705$^*$ & 0.6216$^*$ \\
NoGTO & 0.9527$^*$ & 0.8855$^*$ & 0.7395$^*$ & 0.6428$^*$ \\
HiExpan & 0.9524$^*$ & 0.9045 & 0.7365$^*$ & 0.7132$^*$ \\
Llama-3.1 70B & 0.9708$^*$ & 0.8607$^*$ & 0.8934$^*$ & 0.8010 \\
\midrule
\ours & \textbf{0.9817} & \textbf{0.9210} & \textbf{0.9220} & \textbf{0.8034} \\
\ \ NoParentPretrain & 0.9668$^*$ & 0.7836$^*$ & 0.8920$^*$ & 0.7864 \\
\ \ NoSiblingPretrain & 0.9425$^*$ & 0.9114 & 0.7930$^*$ & 0.6838$^*$ \\
\bottomrule
\end{tabular}
}
\label{tab:hiexpan}
\vspace{-1em}
\end{table}

\vspace{1mm}
\noindent \textbf{Baselines.}
The baselines for seed-guided taxonomy construction include
\textbf{HSetExpan} \cite{shen2017setexpan},
\textbf{HiExpan} \cite{shen2018hiexpan}, two ablation versions of HiExpan---\textbf{NoREPEL} \cite{shen2018hiexpan} and \textbf{NoGTO} \cite{shen2018hiexpan}---as well as \textbf{Llama-3.1 70B} \cite{dubey2024llama}.
Besides, following our practice in the previous two tasks, we consider two ablation variants, \textbf{NoParentPretrain} and \textbf{NoSiblingPretrain}.

\vspace{1mm}
\noindent \textbf{Evaluation Metrics.}
At the top layer, both our \ours model and most baselines achieve near-perfect accuracy. Therefore, our evaluation focuses on the more challenging bottom layer. We use \textbf{Sibling nDCG@$k$} to assess the accuracy of the sibling-finding step and \textbf{Parent nDCG@$k$} to evaluate the accuracy of the parent-finding step.

\vspace{1mm}
\noindent \textbf{Experimental Results.}
Table \ref{tab:hiexpan} demonstrates the Parent and Sibling nDCG@50 scores of compared methods in seed-guided taxonomy construction. We find that:
(1) \ours clearly outperforms all baselines in both the sibling-finding and parent-finding steps across both datasets. Notably, identifying correct sibling terms that are relevant to the taxonomy is a prerequisite for accurately determining their parent categories. If an expanded sibling is incorrect (i.e., it does not belong at this layer or anywhere in the taxonomy), predicting its correct parent becomes impossible. This explains why the Sibling nDCG@50 score is always higher than the corresponding Parent nDCG@50 score.
(2) \ours consistently outperforms the two ablation versions, which is intuitive, as seed-guided taxonomy construction relies on the synergy of both skills.


\subsection{Effect of the LLM Backbone}
Although we use Llama-3 8B as the backbone for \ours in previous experiments, it is important to emphasize that \ours is a versatile framework that can be instantiated with various off-the-shelf generative LLMs. To demonstrate the generalizability of \ours, we evaluate its performance when Llama-2-chat 7B~\cite{touvron2023llama}, Mistral 7B~\cite{jiang2023mistral}, and Gemma 7B~\cite{team2024gemma} are plugged in.

\begin{table}[!t]
\centering
\caption{Performance of \ours with different LLM backbones. For the seed-guided taxonomy construction task (i.e., DBLP and PubMed-CVD), we show Sibling nDCG@50; for the taxonomy expansion task (i.e., Environment and Science), we show Wu\&P.}
\vspace{-0.5em}
\resizebox{\linewidth}{!}{
\begin{tabular}{lcccc}
\toprule
\textbf{Method} & \textbf{DBLP}    & \textbf{PubMed-CVD} & \textbf{Environment} & \textbf{Science} \\
\midrule
Strongest Baseline         & 0.9708            & 0.8934               & 0.777                 & 0.853             \\
\midrule
\ours \\
\ \ Llama-3 8B  & \textbf{0.9817}   & \textbf{0.9220}      & \textbf{0.8300}       & 0.8480            \\
\ \ Llama-2-chat 7B            & 0.9713            & 0.8923               & 0.7739                & 0.7370            \\
\ \ Mistral 7B                 & 0.9635            & 0.9162               & 0.7552                & 0.8437            \\
\ \ Gemma 7B                   & 0.9685            & 0.8627               & 0.7893                & \textbf{0.8713}  \\
\bottomrule
\end{tabular}
}
\label{tab:llm}
\vspace{-1em}
\end{table}

Table~\ref{tab:llm} presents the performance of \ours with different LLM backbones. Due to space limitations, we only display results for 4 datasets (out of the 6 used in the previous experiments) and one metric for each dataset. From Table~\ref{tab:llm}, we observe that:
(1) On DBLP, both Llama-3 8B and Llama-2-chat 7B allow us to outperform the strongest baseline---Llama-3.1 70B, which has a much larger number of parameters; on PubMed-CVD, this could be achieved using Llama-3 8B and Mistral 7B. 
(2) On the Environment dataset, both Llama-3 8B and Gemma 7B enable our framework to beat the best-performing baseline (i.e., TEMP). On the Science dataset, even our default choice Llama-3 8B does not perform the best in Table \ref{tab:hiexpan}, using Gemma 7B allows us to surpass the state of the art.
To summarize, the effectiveness of \ours is built upon the power of our proposed framework and LLMs \textit{in general}, rather than a specific choice of Llama-3 8B.
\section{Related Work}

\noindent \textbf{Entity Set Expansion.} EgoSet~\cite{rong2016egoset} pioneers entity set expansion using skip-grams and word2vec embeddings~\cite{mikolov2013distributed}. Following this, SetExpan~\cite{shen2017setexpan} employs an iterative bootstrapping framework, while CaSE~\cite{yu2019corpus} rank candidates via distributional similarity among context-free embeddings to rank candidate entities according to the seeds.
With pre-trained contextualized language models such as BERT~\cite{devlin2019bert} and GPT-2~\cite{radford2019language}, CGExpan~\cite{zhang2020empower} generates class names to prevent semantic drift, ProbExpan~\cite{li2022contrastive} refines entity representations using contrastive learning, and GAPA~\cite{li2023automatic} leverages autoregressive models for context pattern generation. 
However, all aforementioned approaches do not explore the power of LLMs with billions of parameters and the ability to follow instructions, while \ours extensively exploits the effectiveness of LLMs in entity set expansion.

\vspace{1mm}
\noindent \textbf{Taxonomy Expansion.} Earlier, lexical patterns~\cite{panchenko2016taxi} and distributional word representations~\cite{shwartz2016improving} are used to infer the hypernym-hyponym relationship. 
Later, TaxoExpan~\cite{shen2020taxoexpan} and STEAM~\cite{yu2020steam} propose to encode local ego-graphs and mini-paths, respectively, corresponding to each entity in the taxonomy.
In addition, TMN~\cite{zhang2021taxonomy} examines candidate parents and children via a triplet matching network.
Most recently, TaxoPrompt~\cite{xu2022taxoprompt} and TacoPrompt~\cite{xu2023tacoprompt} adopt prompt tuning on BERT-based encoder models to generate contextualized representations of the global taxonomy structure; BoxTaxo~\cite{jiang2023single} uses box embeddings to replace single-vector embeddings to better capture the hierarchical structure of concepts.
Introducing a more challenging version of taxonomy expansion, \citet{shen2018hiexpan} study seed-guided taxonomy construction which requires the initial step of extracting new entities from text corpora given a small set of seeds before performing taxonomy expansion.
Different from previous approaches that utilize context-free embeddings, graph neural networks, and BERT-based language models, our \ours model unleashes the power of LLMs such as Llama-3. 
More recently, there are studies~\cite{zeng2024chain,zeng2024codetaxo} leveraging GPT-4 and advanced prompting techniques for taxonomy expansion.
By contrast, \ours is a unified framework aiming to jointly solve entity set expansion, taxonomy expansion, and seed-guided taxonomy construction rather than any of them alone.

\vspace{1mm}
\noindent \textbf{Structure-Aware Prompting and Instruction Tuning.} 
There has been increasing attention on prompting and instruction-tuning LLMs to learn from (text-rich) structured data~\cite{jin2023large,li2023survey,chen2024exploring}.
For instance, \citet{wang2023can} strategically prompt LLMs to solve graph problems such as shortest paths and maximum flows;
InstructGLM~\cite{ye2023natural} shows that LLMs fine-tuned on node classification and link prediction can outperform competitive graph neural network baselines;
\citet{zhang2023making} put entity triplets into an instruction template for LLMs to perform knowledge graph completion;
\citet{guo2023gpt4graph} conduct a benchmark study on LLMs' ability to understand graph data by using formal language to describe graphs.
Different from these studies that focus on graph structures (e.g., academic networks), our work specifically explores how taxonomy structures can guide the instruction tuning process to unleash LLMs' potential to solve entity enrichment tasks in a unified way.
\section{Conclusions}
In this paper, we present \ours, a unified framework designed to jointly address entity set expansion, taxonomy expansion, and seed-guided taxonomy construction.
We introduce a taxonomy-guided instruction tuning technique that effectively exploits the existing large-scale taxonomy to teach LLMs the commonality of the three tasks (i.e., the skills of sibling finding and parent finding).
Through extensive experiments on widely used benchmarks for all three tasks, we demonstrate the superiority of \ours over competitive task-specific baselines.

\section*{Limitations}
Our work has the following limitations.
First, since our primary goal is to verify the universal effectiveness of LLM instruction tuning across all three tasks, we intentionally keep our framework as simple as possible, avoiding complex signals utilized in previous studies such as paths~\cite{liu2021temp,jiang2022taxoenrich} and local graphs~\cite{mao2020octet,wang2021enquire}. 
Second, after instruction tuning, \ours can be further equipped with inference-time techniques such as chain-of-thought prompting~\cite{wei2022chain} and self-consistency reasoning~\cite{wang2023selfc}.
Integrating these techniques into \ours could further enhance its performance, which we leave for future work.

\section*{Acknowledgements}
We thank Ming Zhong and Zoey (Sha) Li for their valuable suggestions.
Research was supported in part by US DARPA INCAS Program No. HR0011-21-C0165 and BRIES Program No. HR0011-24-3-0325, National Science Foundation IIS-19-56151, the Molecule Maker Lab Institute: An AI Research Institutes program supported by NSF under Award No. 2019897, and the Institute for Geospatial Understanding through an Integrative Discovery Environment (I-GUIDE) by NSF under Award No. 2118329.


\bibliography{custom}

\appendix
\section{Appendix}
\subsection{Details of Baselines}
For all three tasks, we use Llama-3.1 70B~\cite{dubey2024llama} as one of our baselines, which is directly prompted with the same instructions as \ours. Besides, we consider the following task-specific baselines.

\label{app:baseline}

\subsubsection{Baselines of Entity Set Expansion}
\begin{itemize}[leftmargin=1em]
\item \textbf{EgoSet}~\cite{rong2016egoset} uses skip-gram context features and word2vec embeddings to expand entity sets in multiple facets. 
\item \textbf{SetExpan}~\cite{shen2017setexpan} iteratively selects skip-gram context features from the corpus and proposes a rank ensemble mechanism for scoring and selecting entities.
\item \textbf{SetExpander}~\cite{mamou2018setexpander} learns different text embeddings from different types of context features and trains a classifier to predict whether an entity belongs to a set.
\item \textbf{CaSE}~\cite{yu2019corpus} integrates skip-grams and word2vec embeddings to score and rank entities from the corpus.
\item \textbf{SetCoExpan}~\cite{huang2020guiding} generates auxiliary sets as negative sets that are closely related to the target set and simultaneously co-expand multiple sets.
\item \textbf{CGExpan}~\cite{zhang2020empower} infers the target semantic class names by probing a language model and then utilizes the generated class names to expand new entities.
\item \textbf{SynSetExpan}~\cite{shen2020synsetexpan} jointly conducts two related tasks—synonym discovery and entity set expansion—and utilizes synonym information to enhance expansion performance.
\item \textbf{ProbExpan}~\cite{li2022contrastive} devises an entity-level masked language model with contrastive learning to refine the
representation of entities for entity set expansion.
\end{itemize}

\subsubsection{Baselines of Taxonomy Expansion}
\begin{itemize}[leftmargin=1em]
\item \textbf{TAXI}~\cite{panchenko2016taxi} first extracts hypernym-hyponym pairs from text corpora using substrings and lexico-syntactic patterns, then it organizes the extracted terms into a coherent taxonomy.
\item \textbf{HypeNET}~\cite{shwartz2016improving} employs LSTM to concurrently capture the distributional and relational information between term pairs along dependency paths.
\item \textbf{BERT+MLP}~\cite{devlin2019bert} first acquires term embeddings from a pre-trained BERT model and then inputs the embeddings into a multi-layer perceptron to predict the hypernymy relationship.
\item \textbf{TaxoExpan}~\cite{shen2020taxoexpan} leverages graph neural networks to encode local ego-graphs in the input taxonomy to improve entity representations. In the original paper, context-free word embeddings are used as input features. Following~\cite{yu2020steam}, we replace context-free embeddings with more powerful BERT embeddings for this baseline.
\item \textbf{Arborist}~\cite{manzoor2020expanding} explores heterogeneous edge semantics by employing a large-margin ranking loss to ensure an upper limit on the shortest-path distance between predicted and actual parent nodes.
\item \textbf{Graph2Taxo}~\cite{shang2020taxonomy} utilizes cross-domain graph structures and constraint-based learning of directed acyclic graphs.
\item \textbf{STEAM}~\cite{yu2020steam} learns representations for each pair of (new entity, existing entity) from multiple views using paths sampled from the taxonomy.
\item \textbf{TMN}~\cite{zhang2021taxonomy} proposes a triplet matching network to match a query with hypernym-hyponym pairs. It enables insertion of non-leaf query concepts into an existing taxonomy.
\item \textbf{TEMP}~\cite{liu2021temp} employs pre-trained contextual encoders to predict the position of new concepts by ranking the generated taxonomy-paths.
\item \textbf{GenTaxo}~\cite{zeng2021enhancing} learns the contextual embeddings from their surrounding graph-based and language-based relational information and leverages the corpus for pre-training a concept name generator.
\item \textbf{BoxTaxo}~\cite{jiang2023single} represents entities as boxes to capture their parent-child relationship. It optimizes the box embedding~\cite{vilnis2018probabilistic} of each entity from a joint view of geometry and probability.
\end{itemize}

\subsubsection{Baselines of Seed-Guided Taxonomy Construction}
\begin{itemize}[leftmargin=1em]
\item \textbf{HSetExpan}~\cite{shen2017setexpan} iteratively applies SetExpan at each layer of the input taxonomy. For each expanded bottom-layer node, it uses REPEL~\cite{qu2018weakly}, a weakly supervised relation extraction model, to find the most proper parent at the top layer.
\item \textbf{HiExpan}~\cite{shen2018hiexpan} combines the techniques of flat set expansion, parent-child relationship inference, and global optimization of the taxonomy structure by jointly utilizing skip-grams, context-free text embeddings, and entity types.
\item \textbf{HiExpan-NoREPEL}~\cite{shen2018hiexpan} is an ablation version of HiExpan, which does not utilize REPEL for parent-child relationship inference. Instead, it uses context-free text embeddings only.
\item \textbf{HiExpan-NoGTO}~\cite{shen2018hiexpan} is an ablation version of HiExpan, which does not have the global optimization optimization module.
\end{itemize}
\citet{shen2018hiexpan} have released the output taxonomies\footnote{\url{http://bit.ly/2Jbilte}} of the four baselines above on DBLP and PubMed-CVD, which we use for evaluation.

\subsection{Details of Evaluation Metrics}
\label{app:metric}
\subsubsection{Metric for Entity Set Expansion} 
We use \textbf{MAP@$k$} as the evaluation metric. Formally, given a set of seeds $\mathcal{S}=\{s_1, ..., s_M\}$ and the top-$k$ expanded entities $\mathcal{S}^+=\{s_{M+1}, ...,$ $s_{M+k}\}$, the average precision ${\rm AP}@k$ is defined as 
\begin{equation}
\small
{\rm AP}@k(\mathcal{S}, \mathcal{S}^+) = \frac{1}{k} \sum_{\substack{i: 1 \leq i \leq k, \\ s_{M+i} \sim \mathcal{S}}} \frac{\sum_{j=1}^i \mathbb{I}(s_{M+j} \sim \mathcal{S})}{i}.
\end{equation}
Here, $s_{M+j}$ $\sim$ $\mathcal{S}$ denotes that the expanded entity $s_{M+j}$ and the seed entities in $\mathcal{S}$ belong to the same semantic class; $\mathbb{I}(\cdot)$ is the indicator function. Since there are multiple testing queries (i.e., multiple sets of seeds) $\mathcal{S}_1,...,\mathcal{S}_C$ and their corresponding expansion results $\mathcal{S}_1^+,...,\mathcal{S}_C^+$, the ${\rm MAP}@k$ score is defined as
\begin{equation}
\small
{\rm MAP}@k = \frac{1}{C} \sum_{i=1}^C {\rm AP}@k(\mathcal{S}_i, \mathcal{S}_i^+).
\end{equation}

\subsubsection{Metrics for Taxonomy Expansion}
We use Accuracy (\textbf{Acc}) and Wu \& Palmer Similarity (\textbf{Wu\&P}) as the evaluation metrics.

${\rm Acc}$ is the exact match accuracy of the predicted parent node of each testing entity. Formally, assume the testing set has $C$ samples $x_1, ..., x_C$, and their ground-truth parents in the input taxonomy are $y_1, ..., y_C$, respectively. Then the accuracy of the learned parent-child relationship $\textsc{Parent}^+(\cdot)$ is defined as
\begin{equation}
\small
{\rm Acc} = \frac{1}{C} \sum_{i=1}^C \mathbb{I}(\textsc{Parent}^+(x_i)=y_i).
\end{equation}

${\rm Wu\&P}$~\cite{wu1994verbs} calculates the similarity between the predicted parent and the ground-truth parent based on their distance in the taxonomy.
\begin{equation}
\small
    {\rm Wu\&P} = \frac{1}{C} \sum_{i=1}^C \frac{2\times {\rm depth}({\rm LCP}(\textsc{Parent}^+(x_i), y_i))}{{\rm depth}(\textsc{Parent}^+(x_i)) + {\rm depth}(y_i)},
\end{equation}
where ${\rm LCP}(\cdot, \cdot)$ is the lowest common ancestor of two nodes, and ${\rm depth}(\cdot)$ denotes the depth of a node in the taxonomy.

\subsubsection{Metrics for Seed-Guided Taxonomy Construction}
We use \textbf{Sibling nDCG@$k$} and \textbf{Parent nDCG@$k$} as the evaluation metrics. Formally, in a two-layer taxonomy, given the bottom-layer seeds $\mathcal{S}_2=\{s_{2,1}, ...,$ $s_{2,M}\}$, we examine the top-$k$ expanded bottom-layer entities $\mathcal{S}_2^+=\{s_{2,M+1}, ..., s_{2,M+k}\}$.
${\rm Sibling}\ {\rm nDCG}@k$ evaluates the accuracy of the sibling-finding step (i.e., whether $s_{2,M+i}$ and $\mathcal{S}_2$ belong to the same semantic class).
\begin{equation}
\small
    {\rm Sibling}\ {\rm nDCG}@k = \frac{\sum_{i=1}^k \frac{\mathbb{I}(s_{2,M+i} \sim \mathcal{S}_2)}{\log_2(i+1)}}{\sum_{i=1}^k \frac{1}{\log_2(i+1)}}.
\end{equation}
${\rm Parent}\ {\rm nDCG}@k$ evaluates the accuracy of the parent-finding step. For each expanded bottom-layer entity $s_{2,M+i}$, let $s_{1,p(i)}$ denote its ground-truth parent at the top layer. Then, this metric can be defined as
\begin{equation}
\small
    {\rm Parent}\ {\rm nDCG}@k = \frac{\sum_{i=1}^k \frac{\mathbb{I}(\textsc{Parent}^+(s_{2,M+i}) = s_{1,p(i)})}{\log_2(i+1)}}{\sum_{i=1}^k \frac{1}{\log_2(i+1)}}.
\end{equation}

\section{Hyperparameter Study}
\label{appendix:retriever_u}

\begin{table}[!t]
\centering
\caption{Performance comparison for different values of $U$ (i.e., the number of retrieved candidate parents).}
\renewcommand{\arraystretch}{1.15}
\resizebox{\linewidth}{!}{
\begin{tabular}{c|ccc|ccc}
\toprule
\multirow{2}{*}{$U$} & \multicolumn{3}{c|}{\textbf{Environment}} & \multicolumn{3}{c}{\textbf{Science}} \\
& Llama-3.1-70B & \ours & MaxAcc & Llama-3.1-70B & \ours & MaxAcc \\
\midrule
10  & \textbf{46.15} & 50.00 & 59.62 & 42.35 & 50.59 & 62.35 \\
20  & 36.54 & \textbf{51.15} & 69.23 & 44.71 & \textbf{61.65} & 72.94 \\
40  & 42.31 & 40.38 & 73.08 & 51.76 & 57.65 & 83.53 \\
60  & 44.23 & 44.23 & 80.77 & \textbf{52.94} & 60.00 & 85.88 \\
100 & 34.62 & 40.38 & 84.62 & 49.41 & 60.00 & 90.59 \\
\bottomrule
\end{tabular}
}
\label{tab:retriever_u}
\end{table}

To better understand how the performance of \ours in taxonomy expansion is influenced by the number of candidates retrieved from the taxonomy. We conduct an experiment varying \( U \) (with 10, 20, 40, 60, and 100) and report the results for both Llama-3.1-70B and \ours. Our findings indicate that there is no clear positive correlation between the number of candidate entities in the instruction and the model’s accuracy. A larger \( U \) implies a higher upper bound, as there will be more candidate parent sets that contain the correct parents. This upper bound is represented as MaxAcc in the table. Meanwhile, we also observe that an excessively long context can degrade actual performance. The best choice of \( U \) for Llama-3.1-70B on the Environment dataset is 10, which outperforms the performance at \( U = 100 \) by about 10\%. Additionally, our experiment further confirms the superiority of \ours. Across nearly all values of \( U \), \ours outperforms Llama-3.1-70B, despite the latter being 10 times larger than the backbone of \ours. For \ours, we believe 20 is a generally reasonable choice and adopt it as our default setting.



\end{document}